%% file: root.tex
\documentclass[letterpaper, 10 pt, conference]{ieeeconf}  

\IEEEoverridecommandlockouts                              

\usepackage{hyperref}
\usepackage{url}

\usepackage{booktabs}       
\usepackage{amsfonts}       
\usepackage{nicefrac}       
\usepackage{microtype}      
\usepackage{xcolor}         
\usepackage{kotex}          
\usepackage[edges]{forest}
\usepackage{graphics} 
\usepackage{epsfig} 

\usepackage{times} 
\usepackage{amsmath} 
\usepackage{amssymb}  

\usepackage[
linesnumbered,ruled,vlined]{algorithm2e}
\usepackage{algpseudocode}
\usepackage{algorithmicx}
\usepackage{algcompatible}
\usepackage{threeparttable} 

\usepackage{array}
\usepackage{bm}
\usepackage{graphicx}
\usepackage{subcaption}
\usepackage{caption} 
\usepackage{amsmath}

\usepackage{array}

\newcolumntype{L}[1]{>{\raggedright\let\newline\\\arraybackslash\hspace{0pt}}m{#1}}
\newcolumntype{C}[1]{>{\centering\let\newline\\\arraybackslash\hspace{0pt}}m{#1}}
\newcolumntype{R}[1]{>{\raggedleft\let\newline\\\arraybackslash\hspace{0pt}}m{#1}}

\title{\LARGE \bf
Miniature Testbed for Validating \\ Multi-Agent Cooperative Autonomous Driving
}

\author{Hyunchul Bae$^{\dagger}$, Eunjae Lee$^{\dagger}$, Jehyeop Han$^{\dagger}$, Minhee Kang, \\ Jaehyeon Kim, Junggeun Seo, Minkyun Noh and Heejin Ahn$^{*}$
\thanks{$\dagger$ These authors contributed equally to this work.}
\thanks{$*$ Corresponding author.}
\thanks{Hyunchul Bae, Eunjae Lee, Jehyeop Han, Minhee Kang, and Heejin Ahn are with the School of Electrical Engineering, and Jaehyeon Kim, Junggeun Seo, and Minkyun Noh are with the School of Mechanical Engineering at Korea Advanced Institute of Science and Technology (KAIST), Daejeon, 34141, Republic of Korea.
{\tt\small \{bhc2675, eunjae.lee, jehyeophan, ministop, kjh714, sjg12044, minkyun.noh, heejin.ahn\}@kaist.ac.kr}}%
}

\begin{document}
\maketitle
\thispagestyle{empty}
\pagestyle{empty}

\input{sec/0_abstract}
\input{sec/1_Introduction_v2}

\input{sec/2_related_works}
\input{sec/3_lab_setup}

\input{sec/4_system_architecture_v2}

\input{sec/5_experiments}
\input{sec/6_conclusion}

\addtolength{\textheight}{-0cm}   





\section*{ACKNOWLEDGMENT}
This work was supported by the Institute of Information \& Communications Technology Planning \& Evaluation(IITP)-ITRC(Information Technology Research Center) grant funded by the Korea government(MSIT)(IITP-2026-RS-2023-00259991, 100\%)


\bibliographystyle{IEEETrans}
\bibliography{IEEEabrv, root}



\end{document}

%% file: sec/0_abstract.tex
\begin{abstract}



Cooperative autonomous driving, which extends vehicle autonomy by enabling real-time collaboration between vehicles and smart roadside infrastructure, remains a challenging yet essential problem. However, none of the existing testbeds employ smart infrastructure equipped with sensing, edge computing, and communication capabilities. To address this gap, we design and implement a 1:15-scale miniature testbed, CIVAT, for validating cooperative autonomous driving, consisting of a scaled urban map, autonomous vehicles with onboard sensors, and smart infrastructure. The proposed testbed integrates V2V and V2I communication with the publish-subscribe pattern through a shared Wi-Fi and ROS2 framework, enabling information exchange between vehicles and infrastructure to realize cooperative driving functionality.
As a case study, we validate the system through infrastructure-based perception and intersection management experiments. 
\end{abstract}

%% file: sec/1_Introduction_v2.tex
\section{INTRODUCTION}
\label{sec:introduction}

Autonomous driving technologies have advanced significantly over the past decades, and their widespread deployment is now imminent. However, most progress has focused on stand-alone vehicles, whereas cooperative driving—where vehicles and roadside infrastructure share information and coordinate their actions—remains a challenging but essential problem. The deployment of cooperative driving requires rigorous validation in realistic settings, but such validation continues to face substantial barriers in practice.

To enable such validation, several full-scale testbeds have been established, including M-City~\cite{mcity}, K-City~\cite{kcity}, and AstaZero~\cite{AstaZero}. These facilities provide urban-like environments and diverse traffic scenarios, and are equipped with Vehicle-to-Everything (V2X) infrastructure for connected and automated vehicle testing. While they improve the prospects of commercialization, they also demand extensive space and massive investment, creating a significant entry barrier. For cooperative driving, realistic validation further requires smart infrastructure together with a sufficient fleet of autonomous vehicles to capture multi-agent interactions, which are often prohibitively costly and logistically demanding. As a result, researchers often turn to simulation environments, such as CARLA~\cite{dosovitskiy2017carla}, which provide safer and more cost-effective setups. However, simulations inherently struggle to replicate the physical realities of vehicle dynamics, sensor characteristics, and communication issues.

To bridge the gap between full-scale testbeds and simulations, miniature testbeds have emerged as promising alternatives. These platforms enable controlled experiments at reduced cost and with lower entry barriers. Moreover, prior studies~\cite{brennan1999scaled} derived scaling laws based on Buckingham's $\pi$ theorem and experimentally validated that scaled vehicles exhibit behavior similar to full-scale vehicles. Tab.~\ref{tbl: related work} summarizes representative miniature testbeds, comparing them in terms of vehicle sensing, smart infrastructure (infrastructure with sensing, computation, and communication capabilities), reliance on motion capture systems, and supported interaction types (Vehicle-to-Vehicle (V2V) or Vehicle-to-Infrastructure (V2I)). Most existing platforms rely on onboard vehicle sensors and motion capture systems, while a few explicitly incorporate infrastructure-based sensing or V2I communication. ICAT~\cite{tian2024icat} and MCCT~\cite{dong2023MCCT} introduce infrastructure nodes with V2I communication, but still depend on motion capture systems and infrastructure without sensing capabilities. 


\begin{table}[t!]
\centering
    \scriptsize
    \setlength{\tabcolsep}{2.5pt}
    \caption{Scaled vehicle testing platforms}
    \label{tbl: related work}
    \begin{threeparttable}
    \begin{tabular}{c|ccccc}
    \toprule
    Testbed
    & \begin{tabular}{c}Vehicle\\Sensor\end{tabular}
    & \begin{tabular}{c}Smart\\Infra\end{tabular}
    & \begin{tabular}{c}Motion\\Capture\end{tabular}
    & \begin{tabular}{c}V2V\\Comm.\tnote{1}\end{tabular}
    & \begin{tabular}{c}V2I\\Comm.\tnote{2}\end{tabular}
    \\ \midrule
    \text{Duckietown~\cite{paull2017duckietown}} & \checkmark & -          & -          & -          & -          \\    
    \text{UDSSC~\cite{stager2018delaware}} & \checkmark & - & - & -       & -          \\
    \text{Cambridge~\cite{hyldmar2019cambridge}} & - & -          & \checkmark & -        & \checkmark          \\
    \text{Go-CHART~\cite{kannapiran2020gochart}} & \checkmark & -          & -          & -       & \checkmark          \\
    \text{CPM Lab~\cite{kloock2021cyber}}        & \checkmark & -          & \checkmark  & -  & \checkmark           \\
    \text{ICAT~\cite{tian2024icat}}        & \checkmark & -          & -  &\checkmark  & \checkmark         \\
    \text{MCCT~\cite{dong2023MCCT}}        & \checkmark & -          & \checkmark  & -  & \checkmark          \\
    
    \text{CIVAT}                                 & \checkmark & \checkmark & \checkmark & \checkmark & \checkmark \\
    \bottomrule
    \end{tabular}
    \begin{tablenotes}
        \footnotesize
        \item[1] Direct inter-vehicle communication.
        \item[2] Communication between vehicles and external agents.
    \end{tablenotes}
\end{threeparttable}
\end{table}


In this paper, we propose a novel 1:15-scale miniature testbed, \textbf{CIVAT (Cooperative Intelligent V2X Autonomous Testbed)}, designed for the validation of cooperative autonomous driving technologies. The platform integrates miniature vehicles equipped with onboard sensors and smart infrastructure supported by 3D LiDAR and a powerful edge computing module, thereby enabling advanced perception and decision-making. 
Recognizing that traffic environments will, for the foreseeable future, consist of a mix of connected autonomous vehicles (CAVs) and non-connected vehicles, our testbed supports both fully CAV and such mixed-traffic scenarios, providing a versatile platform for validating cooperative driving algorithms.

\begin{figure*}[t!]
    \centering
    \includegraphics[width=0.98\textwidth]{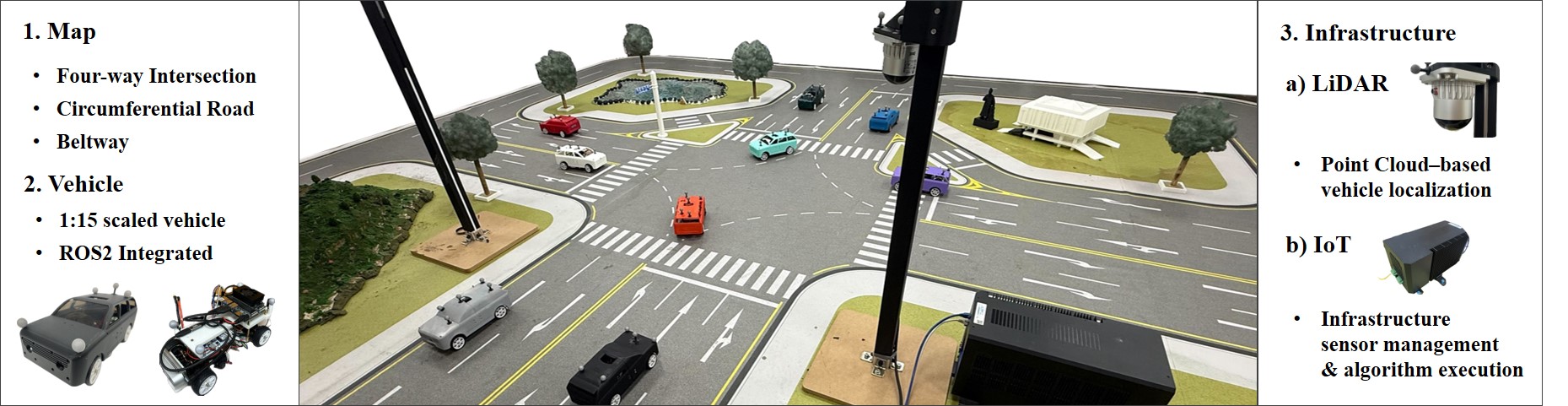}
    \caption{Main components of CIVAT}
    \label{fig_sec3_main_components}
\end{figure*}
The contributions of this study are as follows:
\begin{itemize}
\item We design a 1:15-scale testbed for validating cooperative autonomous driving technologies, incorporating both vehicles and smart infrastructure.
\item We develop miniature vehicles equipped with onboard sensor suites and smart infrastructure equipped with 3D LiDAR and computation devices for perception, computation, and communication.
\item We present an intersection management case study using smart infrastructure's detection with 3D LiDAR and V2I communication to coordinate autonomous vehicles under mixed-traffic conditions.
\end{itemize}

%% file: sec/3_lab_setup.tex
\section{TESTBED DESIGN}
\label{sec:lab_setup}
Our testbed, CIVAT, is implemented by 1:15 scaling down a real-world four-way intersection. As shown in Fig.~\ref{fig_sec3_main_components}, the main components of CIVAT include the map, vehicles, and smart infrastructure. We detail the design of each component and communication in this section.

\subsection{Map Design}
\label{sec:map_design}
\begin{figure}[t]
  \centering
    \includegraphics[width=0.485\textwidth]{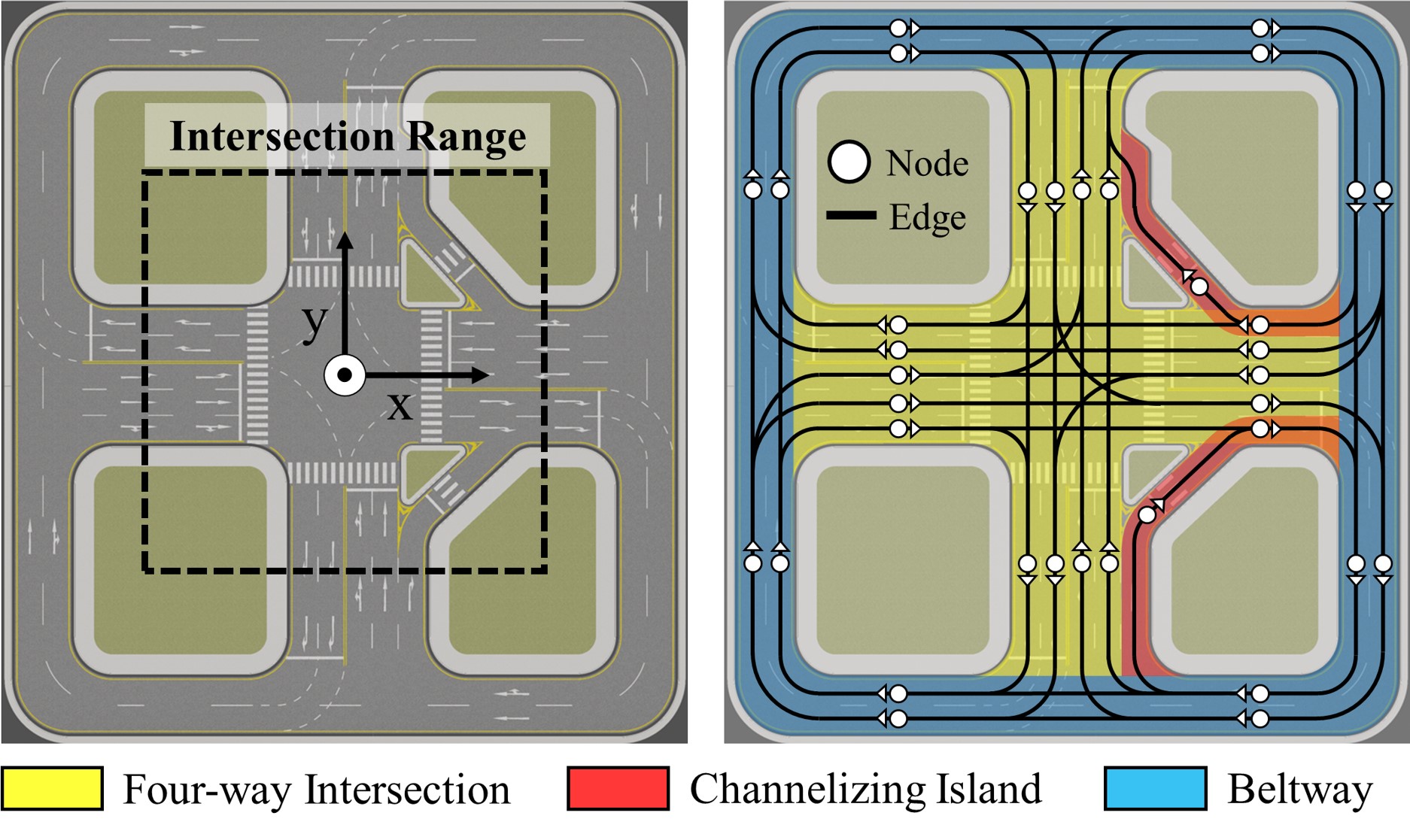}
    \caption{Road types and segments}
    \label{fig_sec3_map_design}
\end{figure}

We design the map shown in Fig.~\ref{fig_sec3_map_design} to capture the complexity of urban traffic and the dynamic interactions between vehicles. We construct it at a 1:15 scale, which yields a physical size of 6\,m~$\times$~5.5\,m corresponding to a real-world scale of 90\,m~$\times$~82.5\,m. At the center, we implement an unsignalized four-way intersection with 2–3 lanes per approach (yellow region in Fig.~\ref{fig_sec3_map_design}) and introduce channelizing islands (red regions in Fig.~\ref{fig_sec3_map_design}) on the approach and departure lanes to induce realistic merging and diverging behaviors. To support continuous vehicle operation, we add a 2-lane beltway (blue regions in Fig.~\ref{fig_sec3_map_design}) that encircles the intersection and serves as a highway for high-speed driving. We model the road network as a graph, where each edge (black lines in Fig.~\ref{fig_sec3_map_design}) corresponds to a path segment. We store the waypoints along every segment in the memory of all agents to enable waypoint following. To replicate the frictional characteristics of asphalt, we print the map surface with material rated at 60 BPN (British Pendulum Number).

\subsection{Vehicle Design}
\label{sec:vehicle_design}
\subsubsection{Hardware Components}

\begin{figure}[t]
  \centering
  \includegraphics[width=0.485\textwidth]{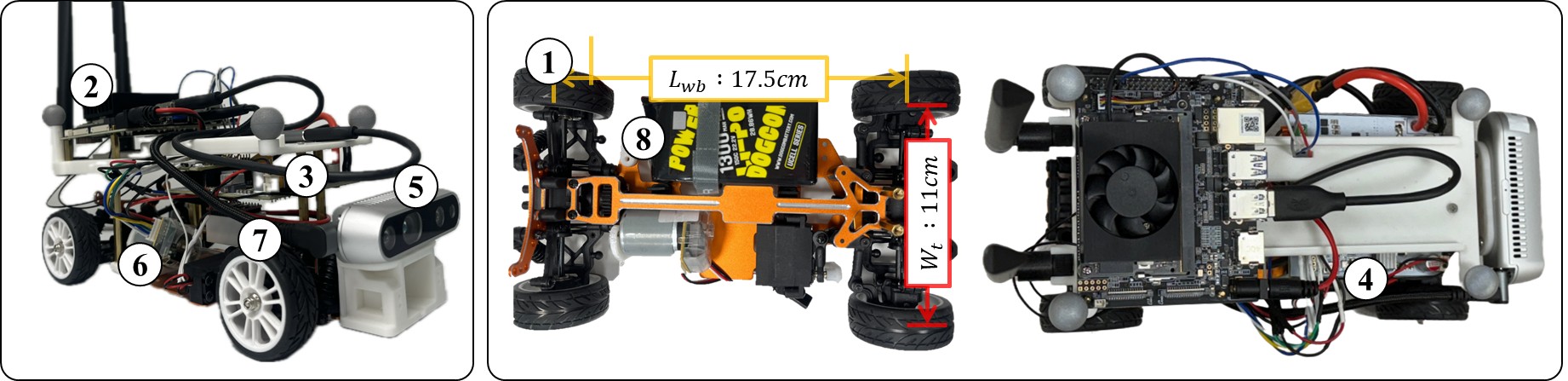}
  \caption{Vehicle components}
  \label{fig_sec3_Vehicle_Component}
\end{figure}

\begin{table}[t!]
\centering
\footnotesize
\setlength{\tabcolsep}{5pt}
\renewcommand{\arraystretch}{0.9}
\begin{threeparttable}
\caption{Hardware components of vehicle}
\label{tab:vehicle_hardware_price}
\begin{tabular}{L{3cm}|L{2.75cm}|L{1.5cm}}
\toprule
\textbf{Component} & \textbf{Item} & \textbf{Price (US\$)} \\
\midrule
1. Chassis \& Servo motor & ZD RACING S16 4WD& 117 \\
2. Single-board Computer & Jetson ORIN NX 16GB & 899 \\
3. Custom PCB & Regulator, Motor Driver & 20 \\
4. Microcontroller Unit & NUCLEO-G431KB & 13 \\
5. Depth Camera & RealSense D435 & 334 \\
6. Encoder DC Motor & M25N-2R-14 12V & 18 \\
7. IMU & myAHRS+ & 76 \\
8. LiPo Battery & Vega 22.2V 1300mAh & 30 \\
\bottomrule
\end{tabular}
\begin{tablenotes}
\footnotesize
\item The number in front of each component corresponds to Fig.~\ref{fig_sec3_Vehicle_Component}. 

\end{tablenotes}
\end{threeparttable}
\end{table}

Fig.~\ref{fig_sec3_Vehicle_Component} represents the overall hardware composition of the vehicle. The vehicle is based on a 1:15-scale chassis\footnote{Available at \url{https://www.alibaba.com/product-detail/ZD-Racing-1-16-2-4G_1600493296884.html}}  with Ackermann steering. It has a wheelbase of 17.5cm and a wheel track of 11cm, and its overall dimensions are $15\,\text{cm}\times30\,\text{cm}\times13\,\text{cm}$.
The onboard system includes the Jetson Orin NX for computation with GPU capabilities, equipped with an external Wi-Fi antenna for V2X communication. It also incorporates a NUCLEO Microcontroller Unit (MCU) for motor velocity control, a RealSense D435 camera for perception, a myAHRS+ IMU for heading angle estimation, and a custom PCB with a power regulator and motor driver. The vehicle is driven by a DC motor with an encoder and is powered by a 1300\,mAh LiPo battery. The total components are summarized in Tab.~\ref{tab:vehicle_hardware_price}, and the total vehicle cost amounts to \$1507.

\begin{figure}[h!]
  \centering
  \includegraphics[width=0.45\textwidth]{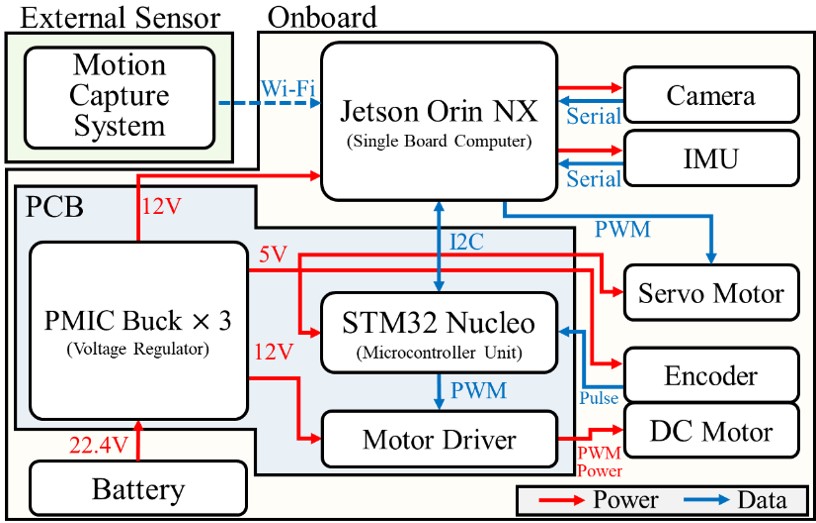}
  \caption{Vehicle system structure}
  \label{fig_sec3_Vehicle_Design}
\end{figure}

Fig.~\ref{fig_sec3_Vehicle_Design} provides an overview of the vehicle system architecture. A single 22.4\,V battery supplies the main power, which is regulated into two 12\,V lines and one 5\,V line via the custom PCB. One 12\,V line powers the Jetson Orin NX, while the other supplies the motor driver. The 5\,V line is used for the MCU (STM32 Nucleo), servo motor, and encoder. Regarding data flow, the Jetson Orin NX serves as the main processing unit on the vehicle. It receives global pose data from the motion capture system via Wi-Fi and sensor measurements from the camera and IMU through serial communication. The Jetson Orin NX sends velocity commands to the MCU via the I2C interface. The MCU converts these commands into PWM signals, which are forwarded to the motor driver to regulate the power supplied to the DC motor. The encoder simultaneously measures the motor rotation and returns pulse signals to the MCU for velocity estimation. In parallel, the Jetson Orin NX issues steering commands directly to the servo motor through PWM signals.

\subsubsection{Kinematic Model}
We model the vehicle with Ackermann steering using the bicycle model~\cite{snider2009automatic}, which captures its essential kinematic behavior under the no-slip assumption:
\vspace{-10pt}
\begin{align}
\begin{split}\label{eq:bicycle_model}
\dot{x} &= v \cos(\psi),  ~~\dot{y} = v \sin(\psi), \\
\dot{\psi} &= \frac{v \tan(\delta)}{L}, ~~ 
\dot{v} = \alpha(v_{ref}-v ), 
\end{split}
\end{align}
where $x$ and $y$ denote the vehicle position, $\psi$ the heading angle, $v$ the longitudinal velocity, $L$ the wheelbase, $\delta$ the steering angle, $v_{ref}$ the reference velocity, and $\alpha$ the inverse time constant. The control inputs are $(v_{ref}, \delta)$.
For the intersection management case study in Section~\ref{sec:experiments}, the smart infrastructure uses this kinematic model to predict future trajectories over a finite horizon $H$, given the current vehicle states $(x, y, \psi, v)$ and their intended paths $P$.

\subsection{Smart Infrastructure Design}
\label{sec:infra_design}

The infrastructure in CIVAT is equipped with an Ouster OSDome LiDAR sensor to capture the surrounding environment and a computation device (ASUS PE6000) that supplies power to the installed sensors and provides computation resources, including a CPU (Intel i9-12900E), a GPU (NVIDIA RTX 4090), and Wi-Fi communication capabilities. The pole is set to 1\,m height to better approximate the real-world infrastructure scale. The detailed components and their prices are summarized in Tab.~\ref{tab:infrastructure_hardware_price}. 
We install the infrastructure near the intersection to perceive and manage objects in a complex urban environment. An infrastructure-based system allows perception and planning to leverage powerful computational resources that are difficult to equip on CAVs, thereby enabling scalability without resource limitations.



To the best of our knowledge, this is the first study to propose infrastructure as an active agent with dedicated sensing and computing modules in a miniature testbed. Prior works have been limited either to V2V communication-based methods~\cite{tian2024icat} or to passive V2I information that provides only fixed guidance, such as traffic lights or road signs~\cite{dong2023MCCT}.



\begin{table}[h!]
\centering
\footnotesize
\setlength{\tabcolsep}{5pt}
\renewcommand{\arraystretch}{0.9}
\caption{Hardware components of infrastructure}
\label{tab:infrastructure_hardware_price}
\begin{tabular}{L{2.5cm}|L{3.7cm}|L{1.3cm}}
\toprule
\textbf{Component} & \textbf{Item} & \textbf{Price (US\$)} \\
\midrule
Computation Device & ASUS PE6000 (i9-12900E)&  2988\\
LiDAR & Ouster OSDome & 7152 \\
Pole & Aluminum Profile 50~$\times$~50 (1m) &  27\\ 
\bottomrule
\end{tabular}
\vspace{-10pt}
\end{table}

\subsection{Communication}
To enable V2V and V2I communication, our testbed establishes a network using IEEE 802.11ax (Wi-Fi 6), interconnecting all vehicles and infrastructure.
While Dedicated Short-Range Communications (DSRC)~\cite{li2010overview} and 3GPP C-V2X~\cite{harounabadi2021v2x} are standardized V2X technologies, they are mutually incompatible and require specialized hardware, whereas Wi-Fi offers broader compatibility, lower cost, and easier deployment. Since our focus is on cooperative driving research that leverages exchanged information between vehicles and infrastructure, we utilize a Wi-Fi-based ROS2 publish-subscribe framework to emulate the message-passing paradigm of V2X. As Fig.~\ref{fig_sec3_map_design} illustrates the V2I communication range, we set the V2I communication range to cover the intersection area rather than the entire testbed, reflecting our focus on intersections where vehicle interactions occur most frequently. In addition, the V2V communication range is capped at 3\,m to emulate realistic communication constraints. 





%% file: sec/4_system_architecture_v2.tex
\begin{figure*}[t!]
  \centering
  \includegraphics[width=0.9\textwidth]{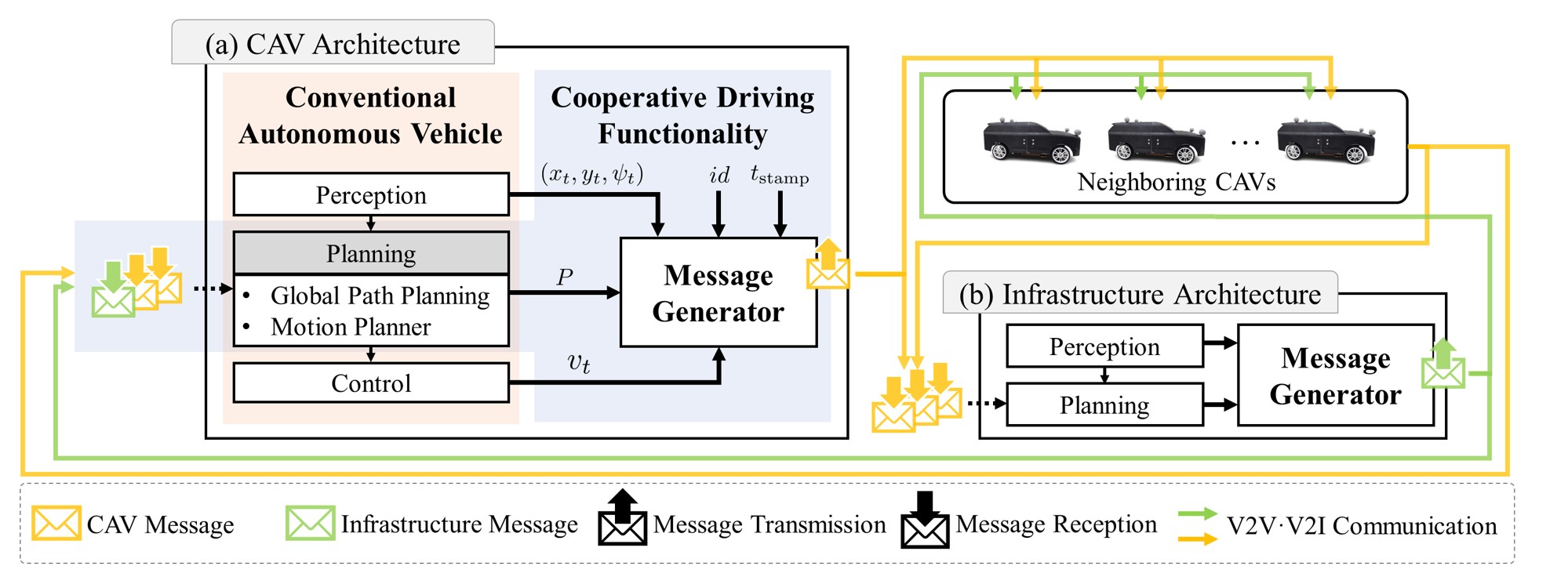}
  \caption{Overall architecture of CIVAT}
  \label{fig_sec4_overall_architecture}
\end{figure*}

\section{Software Architecture}
\label{sec:system_architecture}

We design the software architecture of CIVAT to enable cooperative autonomous driving through both V2V and V2I communication. As shown in Fig.~\ref{fig_sec4_overall_architecture}, all CAVs and the smart infrastructure exchange messages within CIVAT, allowing our testbed to support diverse cooperative driving scenarios.

\subsection{CAV Architecture}


As shown in Fig.~\ref{fig_sec4_overall_architecture}(a), the CAV architecture extend the conventional autonomous driving pipeline by incorporating cooperative driving functionality: each CAV receives messages from other agents (CAVs and infrastructure) and leverages the shared information during the planning stage. At the same time, it transmits its own messages by aggregating the outputs of the perception, planning, and control modules. 

In the following, we detail the planning and message generator modules. Although each vehicle is equipped with onboard sensors for perception, in the current testbed, we employ motion capture measurements to localize each CAV.

\subsubsection{Planning}
\label{sec:motion_planner}

Each vehicle is assigned a destination and generates a global path using the Dijkstra algorithm~\cite{dijkstra2022dijskstra}. The waypoints of the global path, stored in the map, provide the reference for lateral guidance. For longitudinal motion, we adopt the Intelligent Driver Model (IDM)~\cite{treiber2000congested}. The ego vehicle receives messages from neighboring CAVs, identifies the closest CAV on the same path, and adjusts its speed accordingly. Based on this interaction, the planner generates a trajectory that specifies both the path $P$ through the waypoints and the velocity profile along it, thereby preserving safe following distances and ensuring collision-free motion.


\subsubsection{Message Generator}
The message generator is a core module for enabling cooperative driving. Each CAV $n$ collects the 2D pose $(x^n_t, y^n_t, \psi^n_t)$ from the perception module, the current path $P^n$ from the planning module, and the estimated velocity $v^n_{t}$ from the low-level controller using wheel encoder data. It then appends the CAV ID $id^n$ and timestamp $t^n_{\text{stamp}}$. The message is denoted by ${M_t}^n:=(x^n_t, y^n_t, \psi^n_t, P^n, v^n_{t}, id^n, t^n_{\text{stamp}})$ and is published at 10\,Hz. The message set follows the SAE J3216 Class B standard~\cite{SAE_J3216}, which defines the exchange of vehicle driving intents as a standardized approach to enable cooperative driving. 




\subsection{Infrastructure Architecture}
\label{sec:infrastructure_architecture}


While traditional infrastructure relies on passive elements such as traffic lights and signs to manage traffic, our system enables infrastructure to participate actively and dynamically in traffic management. As illustrated in Fig.~\ref{fig_sec4_overall_architecture}(b), we implement the perception and planning modules and broadcast the resulting information to nearby CAVs, thereby enabling V2X cooperative autonomous driving.

\subsubsection{Perception}
We can implement either a stand-alone infrastructure perception module or a collaborative perception module that participates in sensor data sharing through communication. 
Standalone infrastructure perception outperforms vehicle perception in accuracy and robustness to noise, and in collaborative settings, V2X perception surpasses V2V perception by covering the entire intersection more comprehensively~\cite{bae2024rethinkingroleinfra}.
Therefore, V2X cooperative perception that leverages infrastructure is expected to play a central role in future urban autonomous driving.
Collaborative perception tasks include 3D object detection~\cite{bae2024rethinkingroleinfra}, multi-object tracking~\cite{chiu2024dmstrack}, and occupancy prediction~\cite{song2024cohff}.



\subsubsection{Planning}
The infrastructure coordinates multiple vehicles by producing reference trajectories or direct input commands for each CAV based on perception information. Infrastructure-centric planning offers two key advantages. First, it enables globally near-optimal decisions compared to vehicle-based planning, thereby enhancing both safety and traffic efficiency~\cite{lee2026dt}. Also,  it supports computationally demanding planning algorithms, such as learning-based planning~\cite{gu2022stochastic}, which exceed the real-time computational capacity of individual vehicles. Our testbed enables diverse experiments in which infrastructure actively participates, including multi-vehicle coordination and intersection management.



\subsubsection{Message Generator}
The infrastructure exchanges messages with other agents through V2X communication. While we adopt the SAE J2735 standard~\cite{SAE_J2735} as a baseline, existing specifications primarily cover vehicle and traffic signal messages. To the best of our knowledge, there is no standardized message set dedicated to broadcasting infrastructure metadata. In CIVAT, we therefore define a customized infrastructure message set that includes control commands (e.g., $v_{ref}^n$ and/or $\delta^n$), along with the infrastructure ID and timestamp.

\section{Case Study}
\label{sec:case_study}
\begin{figure}[t!]
  \centering
  \includegraphics[width=0.44\textwidth]{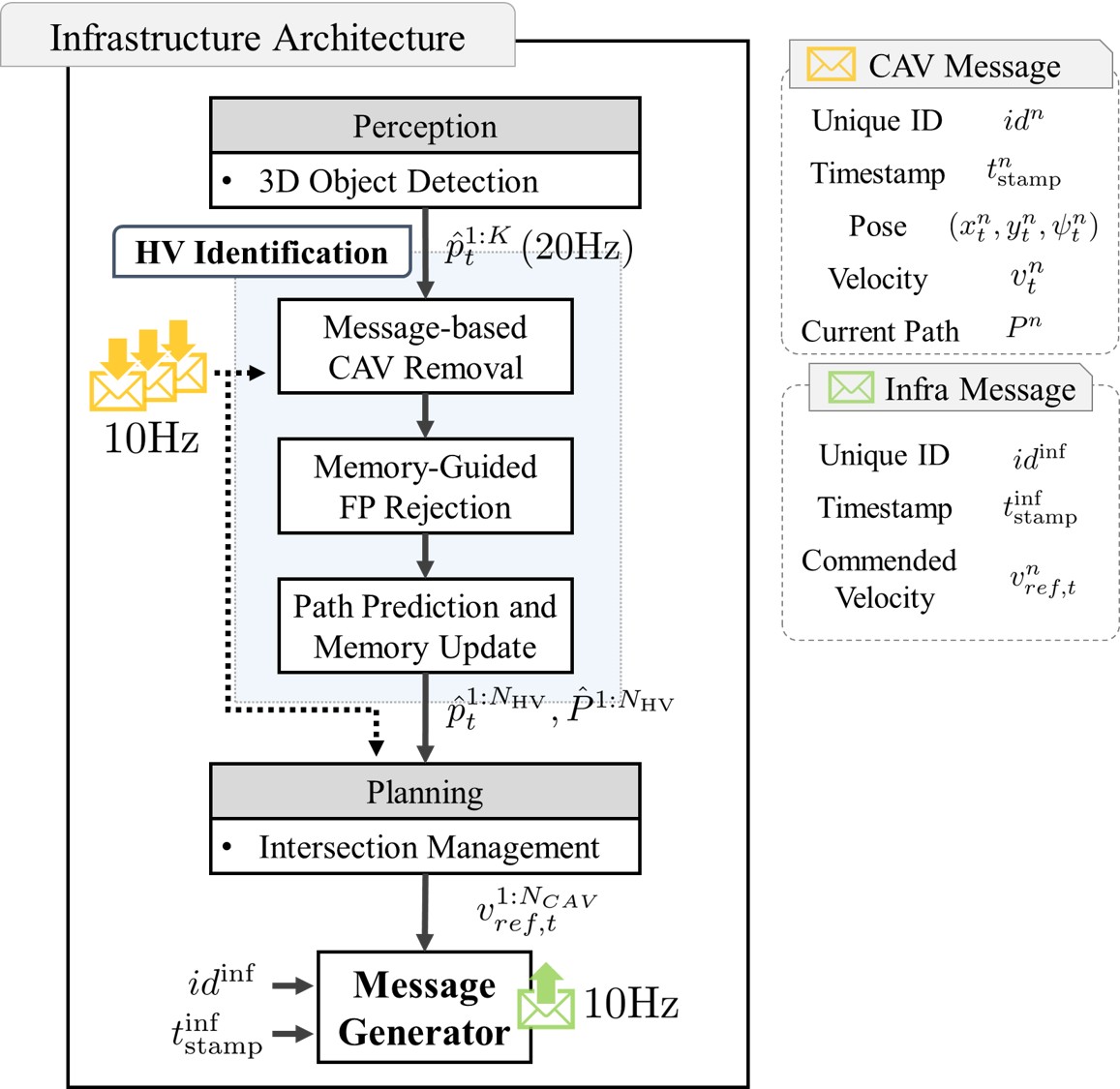}
  \caption{Infrastructure architecture in intersection management}
  \label{fig_sec4_infra_architecture}
  \vspace{-10pt}
\end{figure}

In this paper, we focus on a case study of unsignalized intersection management under mixed-traffic conditions, where CAVs and human-driven vehicles (HVs) coexist on the road. As depicted in Fig.~\ref{fig_sec4_infra_architecture}, the proposed intersection management framework consists of (1) infrastructure-centric 3D object detection, (2) HV identification, (3) an intersection management that determines priority and speed advisories, and (4) a message generator to broadcast both perception outputs and planning decisions to vehicles.



\subsubsection{Infrastructure-centric Object Detection}
The infrastructure must accurately localize all vehicles within the environment. To this end, we integrate an infrastructure-centric 3D object detection module into the intersection management pipeline. We employ PointPillars~\cite{lang2019pointpillars}, which efficiently encodes point clouds into pseudo-images using pillar-wise feature extraction, enabling real-time inference without sacrificing accuracy. We set the LiDAR sampling rate to 20\,Hz and perform object detection by predicting the corner points of 3D bounding boxes. These corner points are then converted into a 2D pose representation consisting of the center coordinates $(\hat{x}^{i}_t, \hat{y}^{i}_t)$ and heading angle $\hat{\psi}^{i}_t$. 
In particular, we only consider bounding boxes that appear within the communication range, shown in Fig.~\ref{fig_sec3_map_design}(a), and exclude bounding boxes outside the region to reflect realistic settings.

In summary, the set of $K$ predicted bounding boxes is denoted by $\hat{p}_t^{1:K} := \{(\hat{x}^{i}_t, \hat{y}^{i}_t, \hat{\psi}^{i}_t)\}_{i=1}^K$ at time $t$. 

\subsubsection{HV Identification}


We implement an algorithm to identify and localize HVs, as they lack communication capabilities. From the object detection results $\hat{p}_t^{1:K}$ and CAV messages ${M_t}^{1:N_{\text{CAV}}}:=(M_t^1, \ldots, M_t^{N_{\text{CAV}}})$ where $N_{\text{CAV}}$ denotes the total number of CAVs at time $t$, we isolate HVs from a mix of CAVs, HVs, and false positives. To achieve this, we design a three-step pipeline summarized in Algorithm~\ref{code:infra-architecture-hv-filter}.

In \textbf{Message-based CAV Filtering}, detected bounding boxes are compared with CAV positions $\{(x^n_t, y^n_t)\}_{n=1}^{N_{\text{CAV}}}$ obtained from messages ${M_t}^{1:N_{\text{CAV}}}$. Bounding boxes within a threshold of $\tau_{\text{CAV}}=13.5$\,cm are removed, as they correspond to CAVs. In \textbf{Memory-Guided FP Rejection}, the remaining bounding boxes are compared with previous HV positions $\{(\hat{x}^n_{t-1}, \hat{y}^n_{t-1})\}_{n=1}^{N'_{\text{HV}}}$, where $N'_{\text{HV}}$ denotes the number of HVs at time $t-1$. Bounding boxes located farther than $\tau_{\text{FP}}=13.5$\,cm from any previous HV position are discarded as false positives. In \textbf{Path Prediction and Memory Update}, the remaining bounding boxes are classified as HVs. Each HV position is matched to nearby entry nodes, and candidate paths are generated from the connected road segments. Paths deviating more than $\tau_\text{p}=20$\,cm from the HV position are eliminated. 

\begin{algorithm}[t!]
\caption{HV Identification Procedure}
\label{code:infra-architecture-hv-filter}

 \textbf{Input:} $\hat{p}_t^{1:K}$, $ M_t^{1:N_{\text{CAV}}}$, $\{(\hat{x}_{t-1}^n,\hat{y}_{t-1}^n)\}^{1:N'_{\text{HV}}}$
 
 \textbf{Output:} $\hat{p}_t^{1:N_{\text{HV}}}, \hat{P}^{1:N_{\text{HV}}}$

 \textbf{Note:} $\mathrm{dist}(\hat{p}_t^i,(x_t^j,y_t^j)):= \sqrt{(\hat{x}_t^i-x^j_t)^2+(\hat{y}^i_t-y^j_t)^2}$

 \textbf{Step 1 — Message-based CAV Filtering}

\For{each detection $\hat{p}_t^i$}{
  \If{$\min\limits_{(x_t^n,y_t^n)}\!\mathrm{dist}(\hat{p}_t^i,(x_t^n,y_t^n))<\tau_{\text{CAV}}$}
    { Remove $\hat{p}_t^i$}
 }

\textbf{Step 2 — Memory-Guided FP Rejection}

\For{each remaining detection $\hat{p}_t^i$}{
  \If{$\min\limits_{(\hat{x}_{t-1}^n,\hat{y}_{t-1}^n)}\!\mathrm{dist}(\hat{p}_t^i,(\hat{x}_{t-1}^n,\hat{y}_{t-1}^n)) \ge \tau_{\text{FP}}$}
     {Remove $\hat{p}_t^i$}
  \Else{
     Keep $\hat{p}_t^i$ as HV}
     }

\textbf{Step 3 — Path Prediction \& Memory Update}

\For{each HV $\hat{p}_t^i$}{
   Find the nearest node and generate candidate paths \\
   Discard paths with $\mathrm{dist}(\hat{p}_t^i,\text{path})>\tau_p$ \\
   Store remaining candidate(s) as $\hat{P}^i$
   }
\textbf{Update memory: } $\hat{p}_t^{1:N_{\text{HV}}},\hat{P}^{1:N_{\text{HV}}}$
\end{algorithm}

\begin{algorithm}[t!]
    \caption{Priority-based Intersection Management}
    \label{code:infra-architecture-intersection-management}
    \textbf{Input: $M_t^{1:N_{\text{CAV}}}$, $\hat{p}_{t}^{1:N_{hv}},\hat{P}^{1:N_{HV}}$}
    
    \textbf{Output: $\{v_{ref,t}^i\}_{i=1}^{N_{\text{CAV}}}$}
    
    \If{CAV in intersection}{
    
    \If{HVs present (identified using \text{Algorithm~\ref{code:infra-architecture-hv-filter})}}{
        Assign the highest priority to HVs \\
        Predict future occupied region using $\hat{p}_{t}^{1:N_{hv}},\hat{P}^{1:N_{HV}},$ and $v_{max}$
        }
    
    \For{each CAV i}{
        Assign priority in entry-time order
        
        $v_{ref, t}^i \gets v_{max}$
        
            \While {$v_{ref, t}^i \geq 0$}{
                Predict CAV $i$'s future occupied region using $M_t^{1:N_{\text{CAV}}}$ and constant $v_{ref, t}^i$
                
                \If{Conflict higher-priority occupied region}{
                    $v_{ref, t}^i \gets v_{ref, t}^i - \Delta v_{step}$ (\textbf{Yield}) 
                }

                \Else{ 
                    \textbf{Break loop} 
                }
            }
    \textbf{Store command: } $\{v_{ref,t}^i\}_{i=1}^{N_{\text{CAV}}}$ 
    }
    }
\end{algorithm}

\subsubsection{Intersection Management}


To coordinate vehicles inside the intersection, the infrastructure regulates each vehicle’s velocity using a priority-based rule, as summarized in Algorithm~\ref{code:infra-architecture-intersection-management}. It first collects neighboring-vehicle information $M_t^{1:N_{\text{CAV}}}$  and HV states $\hat{p}^{1:N_\text{HV}}_t$ and predicted paths $\hat{P}^{1:N_\text{HV}}$ through HV identification. 
To secure the applicability of diverse experiments, we adopt different priority rules depending on the traffic composition: (1) the fully CAV condition, and (2) the mixed-traffic condition.

For the fully CAV condition, we apply a First-In-First-Serve (FIFS) rule as a deterministic conflict resolution mechanism. Vehicles are prioritized by their order of entry into the intersection. The infrastructure then predicts their future occupied regions over the prediction horizon $H$ steps using a kinematic bicycle model~\eqref{eq:bicycle_model}. 
The highest-priority vehicle is assigned the maximum velocity $v_{max}$. For subsequent vehicles, the infrastructure first assigns the maximum velocity $v_\text{max}$ and predicts future trajectories. If a collision is detected, the velocity is iteratively reduced by $\Delta v_{step}$ until a collision-free trajectory is obtained. The resulting value is issued as the control command $v_{ref,t}^i$ to the $i$-th CAV. 

For the mixed-traffic condition, when an HV enters the intersection, it is automatically assigned the highest priority, and its future occupied regions are predicted assuming motion at maximum velocity. If HV identification yields multiple candidate paths, the infrastructure predicts occupied regions for all candidates, and adjusts the commanded velocities of CAVs to ensure they avoid these regions. 




\subsubsection{Message Generator} The message to CAV $n$ includes the unique ID of infrastructure $id^{inf}$ to allow agents to identify the source, a timestamp $t^{inf}_{stamp}$, and the commanded velocity $v_{ref,t}^n$, and is published at 10\,Hz. The 10~Hz update rate is chosen to reflect typical autonomous vehicle communication settings rather than limitations of our system’s computation or scalability. The commanded velocity directly influences the vehicle’s control, enabling efficient and collision-free passage through the intersection.

%% file: sec/5_experiments.tex
\section{Experiments}
\label{sec:experiments}

\subsection{Infrastructure Perception}
\subsubsection{Dataset} For 3D object detection, we construct a dataset collected from standalone infrastructure in CIVAT, referred to as CIVAT-SI. The dataset contains scenarios with up to five vehicles and consists of 3,888 training frames and 1,120 validation frames. Raw sensor data are captured using infrastructure-mounted LiDAR, and ground-truth vehicle poses are obtained from OptiTrack measurements.

\subsubsection{Detection Accuracy} 
\begin{figure}[h!]
  \centering
  \includegraphics[width=0.48\textwidth]{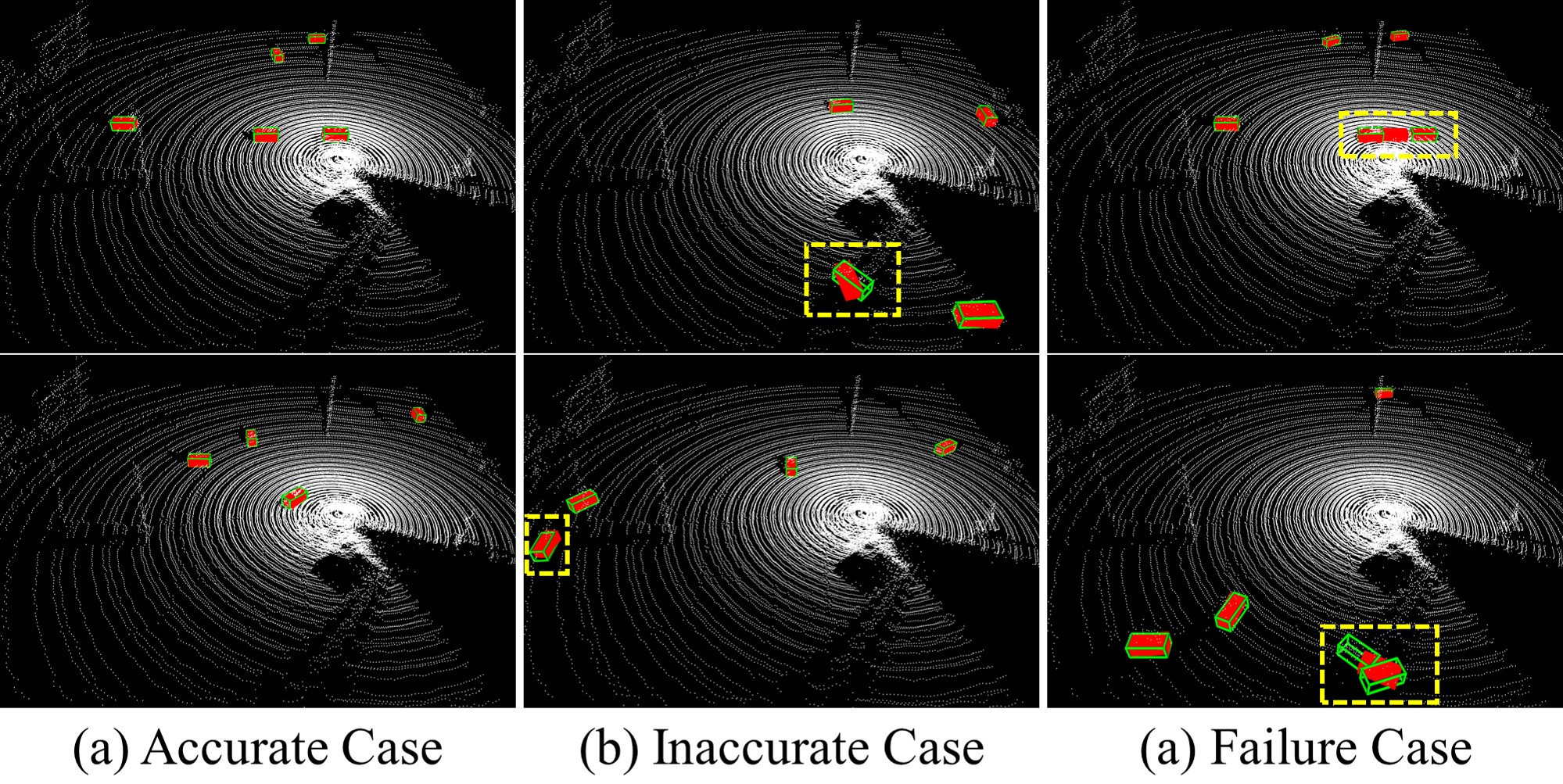}
  \caption{Detection results of infrastructure LiDAR. Green and red boxes denote \textcolor{green}{ground truth} and \textcolor{red}{predictions}, respectively.}
  \label{fig_sec5_detection_results}
\end{figure}

Fig.~\ref{fig_sec5_detection_results} illustrates the 3D object detection results obtained from the infrastructure LiDAR. As shown in Fig.~\ref{fig_sec5_detection_results}(a) and summarized in Tab.~\ref{tab_sec5_detection_results}, most vehicles are detected accurately. Nonetheless, inaccurate or failed detections also occur. In particular, as indicated by the yellow dashed boxes in Fig.~\ref{fig_sec5_detection_results}(b), vehicles located in occluded regions or areas with sparse point clouds often yield inaccurate predictions. Furthermore, as illustrated in the yellow dashed boxes Fig.~\ref{fig_sec5_detection_results}(c), when vehicles are in close proximity, both false positives and false negatives may arise.

\begin{table}[h!]
    \centering
    \caption{Detection accuracy on CIVAT-SI.}
    \label{tab_sec5_detection_results}
    \begin{tabular}{@{}c|ccc}
    \toprule
    Model & AP@0.3 & AP@0.5 & AP@0.7\\ \midrule
    \multicolumn{1}{l|}{PointPillars~\cite{lang2019pointpillars}} & 0.984 & 0.972 & 0.802 \\ \bottomrule 
    \end{tabular}
\vspace{-10pt}
\end{table}

\subsection{Intersection Management}

\begin{figure*}[t!]
  \centering
  \includegraphics[width=0.93\textwidth]{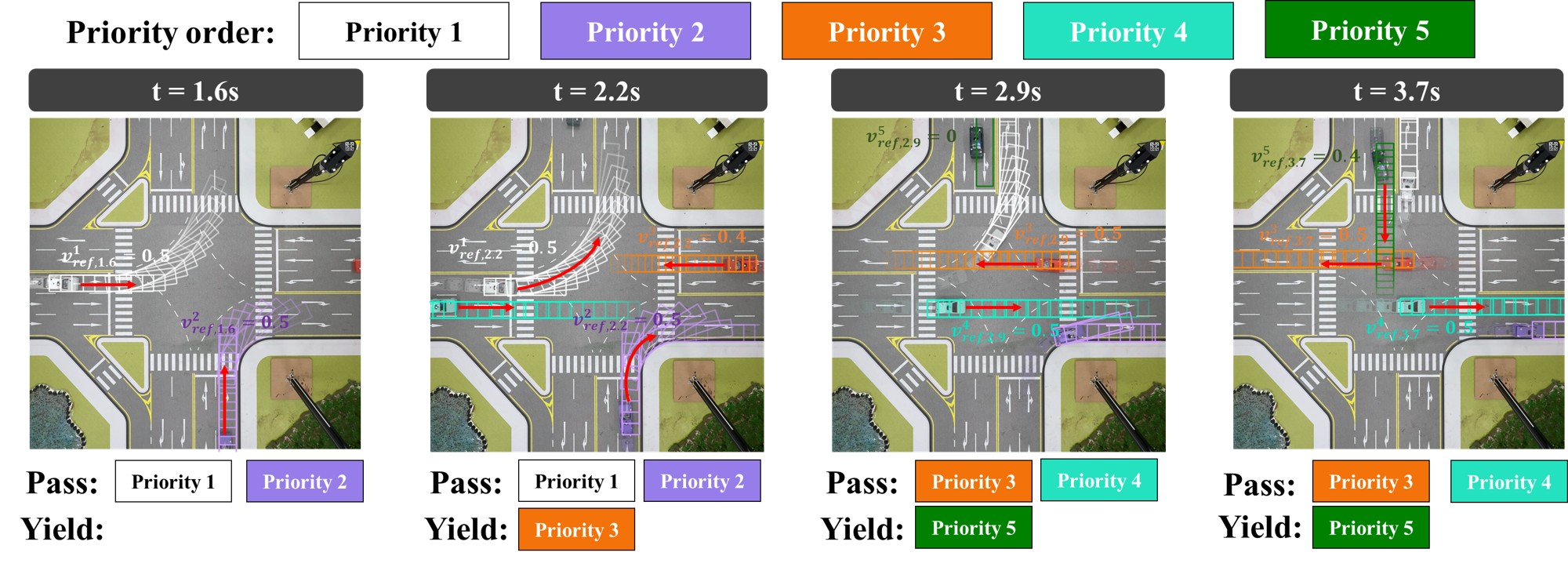}
  \vspace{-8pt}
  \caption{Intersection management results in the fully CAV scenario. Vehicle priorities are indicated at the top. The pass indicates that vehicles approach and pass the intersection with high priority. Yield means that vehicles slow down and halt with low priority. The gradient boxes on the road indicate the occupied regions of the corresponding color-coded agents.}
  \label{fig_sec5_infra_result_CAV}
  \vspace{-10pt}
\end{figure*}

We validate the infrastructure-centric intersection management framework under a fully CAV condition and a mixed-traffic condition. The maximum vehicle velocity is set to $v_{\max} = $0.5\,m/s with a velocity decrement step of $\Delta v_{step} = 0.1$\,m/s. The prediction horizon is set to $H=30$, with a sampling time of $\Delta t = 0.1$\,s. A safety buffer $b_{safe} = $30\,cm is applied when predicting the future occupied region. 

\subsubsection{Fully CAV Condition}
We consider a scenario in which five CAVs enter the intersection, as shown in Fig.~\ref{fig_sec5_infra_result_CAV}.
At $t=1.6$\,s, the first-priority vehicle (white) and the second-priority vehicle (purple) follow non-conflicting paths; thus, the infrastructure assigns both the maximum allowable speed, $v_{ref,1.6}^{1} =v_{ref,1.6}^{2} = 0.5$\,m/s. In contrast, at $t=2.2$\,s, the third-priority (orange) vehicle conflicts with the white vehicle path and therefore yields by decelerating to $ v_{ref, 2.2}^3 = 0.4$\,m/s. At later times ($t=2.9$\,s and $t=3.7$\,s), the infrastructure continues to regulate vehicle velocities according to the priority, ensuring collision-free coordination.


\subsubsection{Mixed-traffic Condition}

\begin{figure*}
  \centering
  \includegraphics[width=0.93\textwidth]{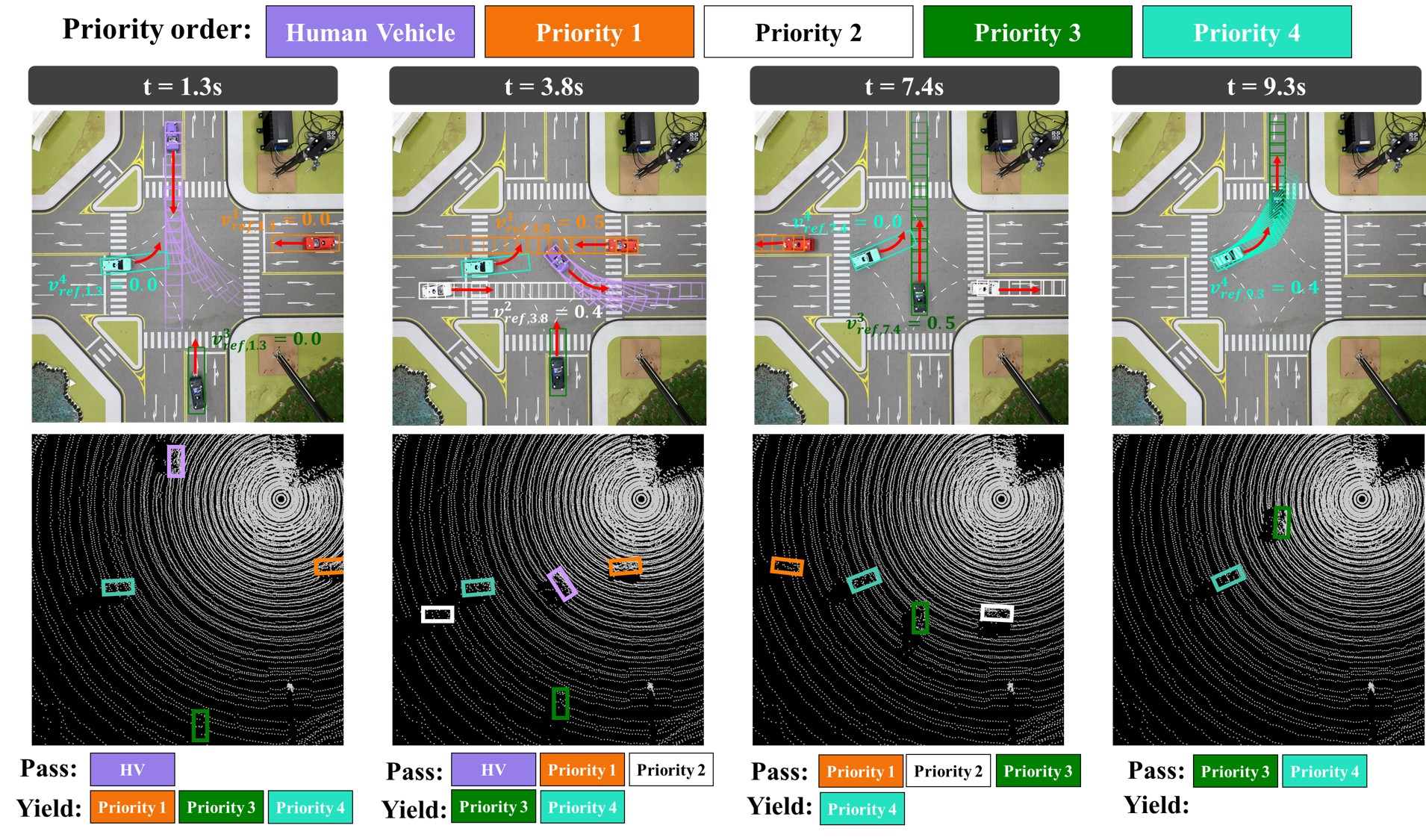}
  \caption{Intersection management and LiDAR detection results in the mixed-traffic scenario.}
  \label{fig_sec5_infra_result_mixed}
  \vspace{-10pt}
\end{figure*}

We evaluate a mixed-traffic condition with four CAVs and one HV entering the intersection concurrently, as shown in Fig.~\ref{fig_sec5_infra_result_mixed}. 
At $t = 1.3$\,s, the HV (purple) has two possible straight or left-turn candidate paths. At this stage, all CAVs stop to yield, allowing the HV to proceed first. By $t = 3.8$\,s, when the HV’s position is closer to the left-turn waypoint and the distance to the straight path exceeds $\tau_\text{p}$, the straight path is discarded and the left-turn path is confirmed as its trajectory. After the HV completes the crossing ($t = 7.4$\,s and $t = 9.3$\,s), the infrastructure resumes priority-based management of the remaining CAVs.

\begin{figure}[t!]
  \centering
  \includegraphics[width=0.45\textwidth]{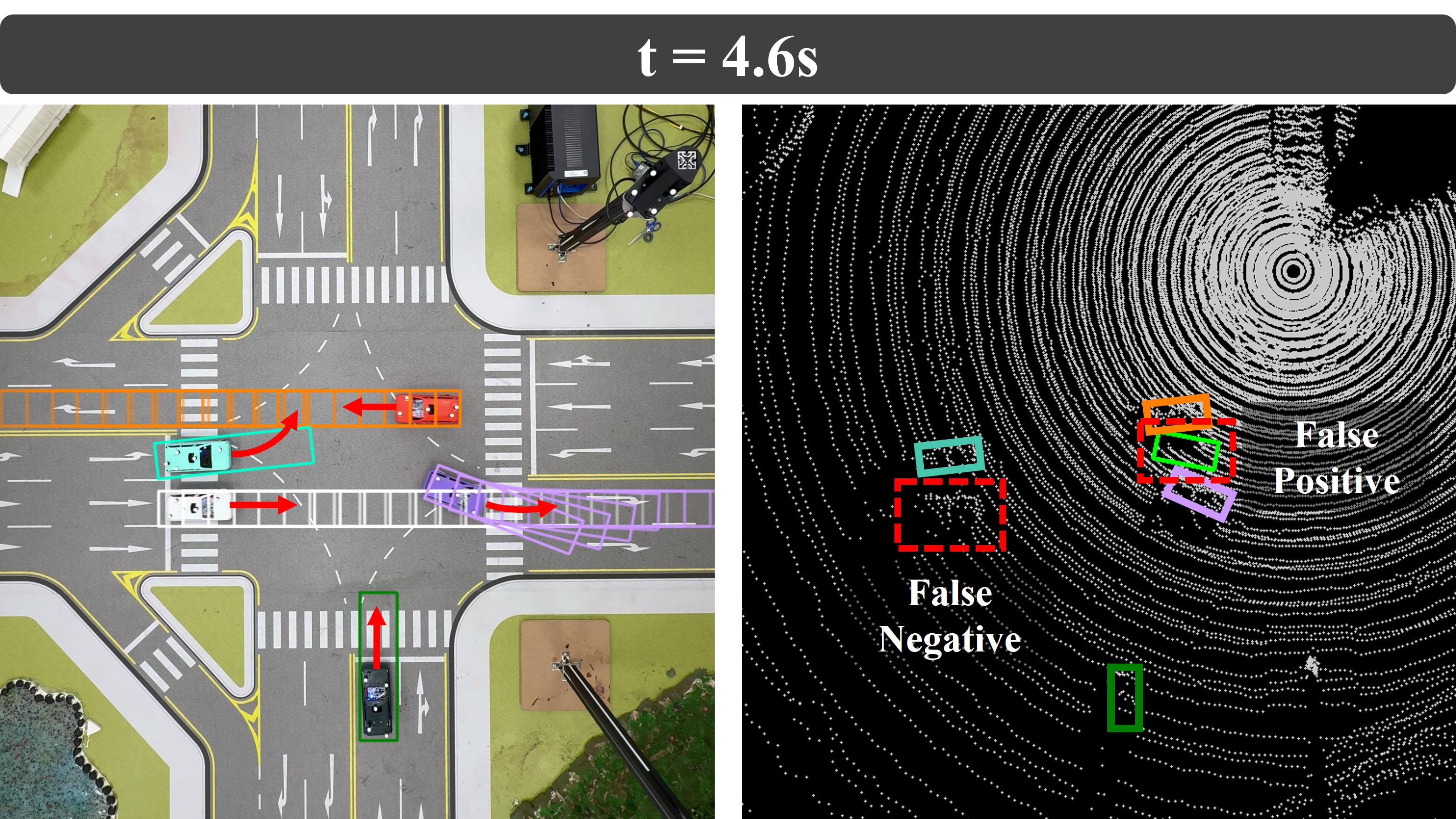}
  \caption{Edge case in mixed-traffic intersection management}
  \label{fig_sec5_mixed_traffic_edge_case}
  \vspace{-10pt}
\end{figure}

As shown in Fig.~\ref{fig_sec5_mixed_traffic_edge_case}, at $t=4.6$\,s the detection results contain both a false positive (FP) and a false negative (FN). The FP is successfully removed by the HV identification algorithm using the previous HV bounding box, while the FN of CAV does not affect management because their positions are determined from the CAV message.
These results highlight the complementary roles of V2I communication and infrastructure perception: (1) V2I communication mitigates the effects of unstable infrastructure-based perception, while (2) perception compensates for the limitations of V2I communication in handling non-communicating agents such as HVs. 

Hence, these experiments show that our testbed highlights the applicability of infrastructure intersection management systems, which have been evaluated mainly in simulations. 

\subsection{Computation Time}
\label{sec:computation_time}
Tab.~\ref{tab_sec5_computation_time} shows the computation times of object detection, HV identification, and intersection management. The computation times are measured from a scenario under the mixed-traffic conditions with four CAVs and one HV.
The average total computation time of the intersection management framework is 31.79\,ms, with a maximum of 40.39\,ms. Given the 100\,ms cycle time of a 10\,Hz system, these results demonstrate that our algorithm satisfies real-time applicability.

As the number of vehicles increases, the computation time of HV identification and intersection management grows due to additional vehicle comparisons and conflict checks. However, these coordination modules are not the computational bottleneck: Tab.~\ref{tab_sec5_computation_time} shows that they require only 0.212~ms on average and 1.140~ms at maximum, whereas object detection alone takes 31.58~ms on average, accounting for 99.3\% of the total runtime.





\begin{table}[h!]
    \centering
    \caption{Computation time by modules (ms)}
    \footnotesize
    \setlength{\tabcolsep}{5pt}
    \renewcommand{\arraystretch}{0.9}
    \label{tab_sec5_computation_time}
    \begin{tabular}{l|ccc}
    \toprule
     \textbf{Module}& \textbf{Min.} & \textbf{Max.} & \textbf{Average}\\ \midrule
    \multicolumn{1}{c|}{Object Detection} & 27.66 & 39.25 & 31.58 \\
    \multicolumn{1}{c|}{HV Identification} & 0.020 & 0.123 & 0.062 \\ 
    \multicolumn{1}{c|}{Intersection Management} & 0.003 & 1.017 & 0.150 \\
    \midrule
    \multicolumn{1}{c|}{\textbf{Total}} & 27.69 & 40.39 & 31.79 \\ \bottomrule 
    \end{tabular}
\vspace{-10pt}
\end{table}


%% file: sec/6_conclusion.tex
\section{Conclusion}
\label{sec:conclusion}

In this paper, we introduced CIVAT, a 1:15-scale testbed for cooperative autonomous driving that integrates miniature vehicles and smart infrastructure. The infrastructure provides 3D LiDAR and computation for perception, communication, and coordination. To demonstrate its utility, we presented an intersection management case study where infrastructure perception and V2I communication coordinate CAVs under mixed-traffic conditions. 
CIVAT provides a cost-effective and scalable platform for evaluating cooperative driving algorithms in both fully connected and mixed-traffic scenarios.

To further demonstrate the utility of CIVAT, we deployed a second instance and hosted the first KAIST Mobility Challenge, where the platform served as an educational testbed for cooperative driving.\footnote{\url{https://www.youtube.com/watch?v=cH-VkOgAqHc}}
 Future work will incorporate onboard sensor-based localization and perception, including an optional SLAM module, to enable end-to-end autonomy evaluation. We will also improve communication realism by emulating V2X impairments (delay, jitter, and packet loss) and potentially integrating standardized V2X communication such as DSRC or C-V2X.


